\documentclass[conference]{IEEEtran}

\newcommand{\etal}{\emph{et~al. \/}}
\newcommand{\given}{$|$}

\def\E{\mathbb{E}}
\def\P{\mathbb{P}}

\def\bf{\bfseries}
\newcommand{\nop}[1]{}

\ifCLASSINFOpdf
  \usepackage[pdftex]{graphicx}
  \DeclareGraphicsExtensions{.pdf,.jpg,.png}      % order is important
  \graphicspath{{.//figs//}}
  
\else
   \usepackage[dvips]{graphicx}
   \graphicspath{{../eps/}}
   \DeclareGraphicsExtensions{.eps}
\fi
\usepackage[cmex10]{amsmath}
\usepackage{algorithmicx}
\usepackage{caption}
\usepackage[font=footnotesize]{subfig}
\usepackage{url}
\usepackage[font={small,it}]{caption}
\usepackage{float}
\usepackage{subfig}
\usepackage{hyperref}
\usepackage{amssymb}
\usepackage[vlined,ruled]{algorithm2e}
\usepackage{tikz}
\usepackage{amssymb} 
\usepackage{multirow}

\IEEEoverridecommandlockouts

\begin{document}

% paper title - needs something more catchy/relating to analysis
\title{On the Veracity of Cyber Intrusion Alerts Synthesized by Generative Adversarial Networks
\vspace*{-20pt}}

\author{\IEEEauthorblockN{
Christopher Sweet, Stephen Moskal, Shanchieh Jay Yang\thanks{This research is supported by NSF SaTC Award \#1526383.}
}

\IEEEauthorblockA{Department of Computer Engineering \\
Rochester Institute of Technology \\
Rochester, New York 14623 \\ Email: Jay.Yang@rit.edu}
}

\maketitle

\begin{abstract}
% Abstract rule-of-thumb: 1-sentence to describe the problem to be solved, 1 sentence to contrast existing works at an extremely high level, 1-2 sentences to describe the novelty of the approach, and 1-2 sentences to highlight the main findings.

Recreating cyber-attack alert data with a high level of fidelity is challenging due to the intricate interaction between features, non-homogeneity of alerts, and potential for rare yet critical samples. Generative Adversarial Networks (GANs) have been shown to effectively learn complex data distributions with the intent of creating increasingly realistic data. This paper presents the application of GANs to cyber-attack alert data and shows that GANs not only successfully learn to generate realistic alerts, but also reveal feature dependencies within alerts. This is accomplished by reviewing the intersection of histograms for varying alert-feature combinations between the ground truth and generated datsets. Traditional statistical metrics, such as conditional and joint entropy, are also employed to verify the accuracy of these dependencies. Finally, it is shown that a Mutual Information constraint on the network can be used to increase the generation of low probability, critical, alert values. By mapping alerts to a set of attack stages it is shown that the output of these low probability alerts has a direct contextual meaning for Cyber Security analysts. Overall, this work provides the basis for generating new cyber intrusion alerts and provides evidence that synthesized alerts emulate critical dependencies from the source dataset.

\end{abstract}


\IEEEpeerreviewmaketitle

\vspace*{-5pt}

\section{Introduction}
\label{sec:intro}

Classifying, predicting, and generating cyber-attack alert data provides a unique set of challenges due to imbalance and a lack of homogeneity in alert datasets. Furthering these challenges critical exploits in a network are often rare and difficult to identify. Despite this is has been shown that alert data can be used to identify anomalous traffic \cite{Veeramachaneni2016} \cite{Filonov2017} \cite{Filonov2016}, network vulnerabilities \cite{Noel2009}, and bad actor behavior profiling \cite{Fava2008}. However, to fully realize the potential of cyber-attack alert data, a means to acquire more data and analyze critical dependencies within alerts is needed. 

This work seeks to provide solutions to these challenges by showing that deep learning models are able to recreate cyber-attack alert data when given representative real world data. This includes a means for driving better coverage of the feature domain in model outputs, allowing more rare but critical events to be synthesized. 

First proposed by Goodfellow \etal, Generative Adversarial Networks (GANs) \cite{Goodfellow2014} are a semisupervised deep learning model that learn to emulate data from a ground truth dataset. This framework was subsequently improved by Arjovsky \etal \cite{Arjovsky2017} and Gulrajani \etal \cite{Gulrajani2017} by optimizing models via the Earthmover Distance. Finally, Belghazi \etal \cite{Belghazi2018} introduced an additional loss term to drive diverse model outputs. Through these improvements, GANs have achieved state of the art results in generating data with respect to images \cite {Karras2018} \cite{Zhu2017} \cite{Ledig2016}, text \cite{Su2018}, and sound \cite{Dong2018} \cite{Gao2018}. 

Additionally, GANs have been applied to network traffic to modify and obfuscate malicious traffic \cite{Rigaki2018} \cite{Lin2018} \cite{Hu2017} \cite{Anderson2017}. These adversarial samples are created to avoid being flagged by Network Intrusion Detection Systems (NIDS). However, despite the usage of GANs to create data in other fields and modify alerts for malicious intent in Cyber Security, there is a lack of research in using GANs to synthesize NIDS alerts for analysis from the \emph{target} IP perspective. Given the low signal to noise ratio for malicious alerts, a means to generate new data based off historical data could enable researchers to better understand network vulnerabilities and potential attack paths. 

This research applies GANs to NIDS data collected via Suricata (\texttt{https://suricata-ids.org/}) from the 2017 and 2018 Collegiate Penetration Testing Competition (CPTC - \texttt{https://nationalcptc.org/}).  CPTC'17 had ten student teams attempt to penetrate and exploit vulnerabilities of a virtualized network that manages election systems. CPTC'18 tasked new student teams with penetration into an autonomous driving framework including virtualized embedded systems, mobile applications, and processing servers. Rather than directly focusing on the specific behaviors exhibited by each team for the two datasets, the data is segmented based off the IP address being attacked. Doing so allows for alert introspection on a \emph{target by target} basis when comparing ground truth data to that synthesized by a GAN. 

Statistical analysis of these samples is shown to reveal the challenge of generating cyber-attack alert data and natural dependencies between alert features. For example, can the alert signature and destination port category be used to identify when in the competition the alert occurred? And to what degree of accuracy?  These dependencies may then be confirmed through direct computation of conditional entropy. By showing GAN learns these intra-alert feature dependencies we show that GAN is successful at learning latent interactions between feature values when data is limited, non stationary, and lacks homogeneity.

The remainder of the paper is structured as follows: Section \ref{sec:related} provides an overview of some of the existing challenges in Machine Learning for cyber-security as well as existing applications of GANs for Cyber Security data. Section \ref{sec:prob} and Section \ref{sec:experiments} discusses the GAN model as well as preprocessing and analysis methods employed for generating synthetic intrusion alerts. Finally, Section \ref{sec:results} discusses the observations made from reviewing generated data and Section \ref{sec:conclusion} gives the concluding remarks and future works of this research.

\section{Related Work}
\label{sec:related}

With the regularity and complexity of cyber attacks increasing, so has the interest in applying Machine Learning techniques to classify and predict attack actions. However, with the use of Deep Learning models comes the need for massive amounts of quality training data; several ongoing works in this field cite the need for more data as a limitation to their current research \cite{us} \cite{Shen2018} \cite{Faber2018} \cite{Amit2019}. 

In particular, LSTM models are shown by Perry \etal \cite{us} to suffer significantly lower accuracy when the dataset provided for training is not large enough to be representative of previous observations. This holds true for both classifying cyber attackers and for predicting the next attack action taken.

This message is echoed by Faber and Malloy \cite{Faber2018} despite having a dataset of over 600,000 alerts and promising classification accuracy. They note that the availability of quality labeled data and a low signal to noise ratio for malicious activity are both outstanding issues. 

Another avenue for research applying Machine Learning to cyber-security data has been the generation of adversarial traffic. Specifically, GANs have been used to obfuscate malicious traffic through the modification of packet behavior. 

Rigaki \etal \cite{Rigaki2018} proposed the use of GANs in generating network traffic which mimics other types of network traffic. In particular, real malware traffic was modified by a GAN to appear as legitimate network traffic. This allowed the malware to avoid detection from the Stratosphere Behavioral Intrusion Prevention System through the modification of three network traffic parameters; total byte size, duration of network flow, and time delta between current network flow and the last network flow. They showed that through the modification of these parameters detection rate could be dropped down to 0\%. 

Similarly, Lin \etal \cite{Lin2018} apply GANs to obfuscate traffic with the intention of directly deceiving a NIDS. Their model makes use of 9 discrete features and 32 continuous features to modify attack actions to avoid detection. Available attack actions include denial of service and privilege escalation. Their model is shown to drastically increase the evasion rate of malicious network traffic across several different classifiers when benchmarked using the NSL-KDD benchmark provided by \cite{Hu2015}.

Despite the successes of these works, no current GAN model has been applied to recreation or expansion of cyber-attack alert data from the target perspective. This research aims at recreating malicious NIDS samples from the target perspective to expand and better understand attack actions taken against each machine in a network.  

Applying GANs to Cyber Intrusion Alerts is non trivial as the challenges posed by the data directly affect GANs. The distribution of alert features cannot be modeled trivially and critical alert features may occur with low probability. The potential for mode dropping is simultaneously high and problematic to the contextual meaning of results. In order to try and address this in other fields Belghazi \etal \cite{Belghazi2018} propose adding a Mutual Information constraint on the Generator. Applied to Cyber Intrusion Alerts, the Mutual Information constraint would encourage the generation of all alert features, including the rare actions that are indicative of targeted attacker behavior.

Training on a per target IP basis from the CPTC'17 and CPTC'18 datasets shows the promise of GAN as a means to synthesize alerts with limited and non-stationary data. Additionally, intra-alert feature dependencies are captured and revealed by the data sampled from the GANs' output, showing that critical interactions between feature values are preserved by GANs. Finally, using a mapping of alert signatures to believed attack stages shows that the generated results have a direct contextual meaning for cyber analysts.  

\section{GAN Model}
\label{sec:prob}

A Generative Adversarial Network is a class of neural network where two neural networks are pitted against each other. One network, the generator (G), attempts to create samples which seem to belong to a ground truth dataset. The other network, the discriminator (D), takes inputs from the ground truth dataset as well as G, and flags samples as either real or fake. This structure minimizes the generator loss each time G successfully generates a sample that tricks D into marking the sample as real. Conversely, the discriminator loss is minimized when all samples from the ground truth set are marked as real and all samples created by G are marked as fake. 

The Wasserstein GAN, first proposed by Arjovsky \etal \cite{Arjovsky2017} extends the concept of a GAN but with increased stability during training. This was subsequently improved by Gulrajani \etal \cite{Gulrajani2017} by adding a gradient penalty term to regularize the gradients of D. The gradient penalty creates a 1-Lipschitz constraint on the discriminator during training by sampling noise from $\P_z$ and constraining the gradient of the L2 norm of D($\P_z$) to 1.  Additionally, D is given real samples $\P_r$ and generated samples $\P_g$ in a 5:1 ratio per epoch of training; this is done to increase the utility of gradients provided by D. These modifications resulted in the discriminator loss function provided in (\ref{eq:disc_loss}). This model is referred to as Wasserstein GAN with Gradient Penalty (WGAN-GP).

\begin{equation}
    D_{Loss} = \underbrace{\E[D(\P_r)] - \E[D(\P_g)]}_\text{Wasserstein Distance} + \underbrace{\lambda \E[(||\nabla_{\widehat{x}} D(\P_z)||_2-1)^2]}_\text{Gradient Penalty}
    \label{eq:disc_loss}
\end{equation}

Despite these improvements to the loss function for the discriminator, the generator loss was left unmodified. Belghazi \etal \cite{Belghazi2018} changed this by adding a mutual information term to the generator's loss. This contribution maximized an approximation of the mutual information between the generator's noise input $\P_{z}$ and it's output samples $\P_{g}$ by minimizing the Donsker-Varadhan (DV) representation of the Kullback-Leibler (KL) divergence. This modification is shown in (\ref{eq:gen_loss}). 

The DV KL term was added by using a neural network to learn how to estimate Mutual Information between two distributions. The rationale behind this added constraint was that it would force the generator to further explore the domain of the data when generating new samples; not exploring the dataset would result in a limit to the amount of mutual information which could be found between input noise and the output samples. Herein this model will be referred to as the WGAN-GPMI model. 

\begin{equation}
    G_{Loss} = \underbrace{-\E[D(\P_g)]}_\text{Adversarial Loss} + \underbrace{\E[\P_{gz}] + \log(\E[e^{\P_{g} \otimes \P_{z}}])}_\text{DV KL Divergence}
    \label{eq:gen_loss}
\end{equation}

Since mutual information is theoretically unbounded, gradient updates resulting from it could overwhelm the adversarial gradients resulting from the WGAN-GP's discriminator loss function. In order to address this all of the gradient updates to the generator were adaptively clipped to ensure that the Frobenius norm of the gradient resulting from the mutual information was at most equal to the adversarial gradient \cite{Belghazi2018}, as shown in (\ref{eq:fnorm}).

\begin{equation}
	g_{norm} = g_a + min(||{g_a}||, ||{g_m}||)({g_m \over ||g_m||})
	\label{eq:fnorm}
\end{equation}
 
Note that $g_{norm}$ is the normalized gradient, $g_a$ is the adversarial gradient resulting from (\ref{eq:disc_loss}), and $g_m$ is the gradient resulting from the DV KL portion of  (\ref{eq:gen_loss}). 

A WGAN-GP and WGAN-GPMI model were implemented to generate cyber intrusion alerts from historical data. Given the propensity for the WGAN-GPMI model to explore the feature space it should follow that the WGAN-GPMI model will produce superior results to the WGAN model. Each model was configured with a hidden dimension size of $128$, batch size of $100$, and learning rate of $5e-4$. G was configured to sample 64 points of noise per sample batch. Both G and D were two-layer fully connected neural networks. 

\begin{figure}[!htbp]
    \centering
    \includegraphics[width=\columnwidth]{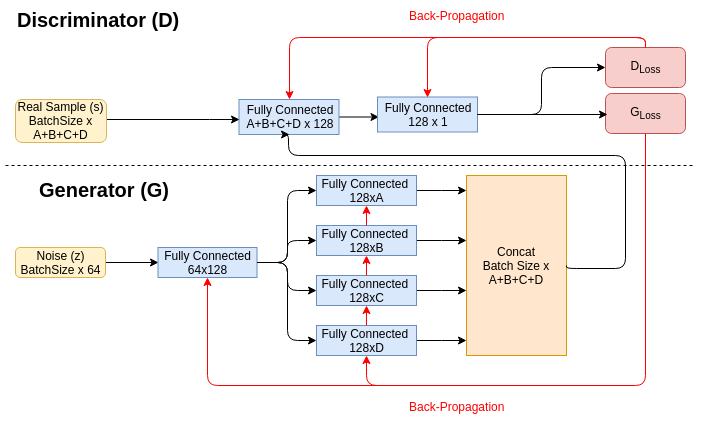}
    \caption{WGAN-GP Model Architecture: The real samples provided to the Discriminator are one hot encoded in the same fashion as the Generator's output}
    \label{fig:gan_sys}
\end{figure}

\begin{figure}[!htbp]
    \centering
\includegraphics[width=\columnwidth]{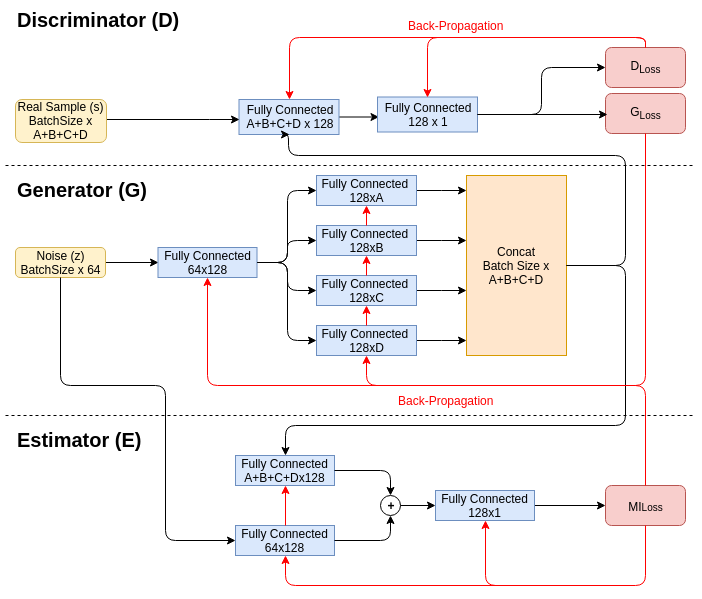}
    \caption{WGAN-GPMI Model Architecture: The Generator and Discriminator are left unmodified from the WGAN-GP model}
    \label{fig:model_complex}
\end{figure}

The generator featured 4 independent fully connected layers in parallel on the output. These generate each of the 4 features tested. The mutual information estimator network also consisted of 2 layers. The first layer took input from each of the aforementioned sources and mapped them to separate hidden representation layers and added together. Then the second layer mapped the hidden representation to a single output value representing the mutual information estimate. The network architecture for WGAN-GP may be seen visually in Fig. \ref{fig:gan_sys} and the addition of the estimator in WGAN-GPMI may be seen in Fig. \ref{fig:model_complex}.

Due to the categorical nature of the data being generated all features were one hot encoded and concatenated into a single vector per alert. These values were then transformed into real-world values by segmenting the vector into subcomponents whose length's equal the number of unique values for the given feature. The argmax of each of these subcomponents was then taken as a post-processing step to find the corresponding real world value generated. These discrete values were utilized for all analytics applied in Section \ref{sec:as_res}.

In both figures yellow boxes represent inputs to the network. The blue boxes represent weight layers of the network which are updated via back-propagation. The concatenation box at the end of the generator is a post processing step to form the aforementioned one hot encoded alert vector from each of the feature outputs. And the red boxes and lines represent feedback paths which update network parameters during each step of training.

\section{Design of Experiments}
\label{sec:experiments}

Training and testing of the CPTC dataset was broken up into four stages. First, a GAN was trained to learn the distribution of the input data on a per target IP basis and emulate it. Then the intersection of histograms was calculated for all combinations of features to express how well the GAN had learned to emulate the dataset and identify feature dependencies. Next, feature dependencies for varying numbers of feature permutations were verified using the weighted, normalized, conditional entropy. Finally, the number of output modes dropped for each model was compared to show that the WGAN-GPMI model covered a larger percentage of the alert feature domain. Additionally, we show that the additional output modes captured by the WGAN-GPMI model pertain to attack stages that the WGAN-GP model fails to capture. 

\subsection{CPTC Dataset \& Preprocessing}
\label{sec:dataset}

The data used for these experiments comes from the National Collegiate Penetration Testing Competition from 2017 and 2018. In 2017, teams were tasked with penetrating and exploiting network vulnerabilities in a virtualized network managing election systems. In 2018 teams were required to attack a multifaceted system handling autonomous cars which included host based systems, servers, and mobile assets such as cell phones running an app. Each team had around 9 hours to scan, infiltrate the network, and exfiltrate information from the target. Both datasets provide a unique opportunity for Machine Learning experimentation as they are completely comprised of malicious actions as teams attempt to compromise the target network. 

Prior to being input to the models as training data, significant preprocessing was done. This not only reduced the dimensionality of each of the features but also increased the contextual utility of generated alerts. Though this data is unique to the competition it is worth noting that the preprocessing described herein is applicable to any dataset consisting of NIDS alerts.

The first preprocessing step applied to the data was to separate alerts on a per-Destination IP basis. This allowed individual models to be trained for each system on the network, typifying the type of traffic seen at that target. Additionally, data from all of the teams could be compounded, more unique attacker behaviors to be captured for each target. 

Segmentation on a per-target basis has several intuitive benefits: First, it allows for different vulnerabilities to be highlighted on each machine given commonly occurring alert features at that target. Secondly, it helps to remove noisy alert influence from critical nodes in the network. For example, internet facing IPs may contain a significant amount of scanning activity, drowning out exfiltration related alert features at nodes further embedded in the network. Finally, the information extracted from alerts on a per target basis is actionable, as network administrators can use commonly targeted vulnerabilities to tune network settings for future defense. Table \ref{table:mapping} shows 4 target IP addresses selected for experimentation from the CPTC'17 dataset as an example; the operating system and high-level purpose of the machine is also given.

\begin{table}[!htbp]
    \caption{CPTC 2017 Mapping of IP Address to Machine Usage/Purpose}
    \label{table:mapping}
    \centering
    \begin{tabular}{c|c|c}
        \textbf{IP Address} & \textbf{Operating System} & \textbf{Machine Usage} \\
        \hline 
        10.0.0.100 & Windows & Active Directory Server \\
        \hline
        10.0.0.27 & Ubuntu & HTTP Server \\
        \hline
        10.0.0.22 & Ubuntu & MySQL Server \\
        \hline
        10.0.99.143 & Ubuntu & HTTP Server
    \end{tabular}
\end{table}

Next, the dimensionality of the destination port feature was reduced based off common service categories run across a collection of ports provided by the Internet Assigned Numbers Authority \cite{iana}. This reduction drops the number of unique values from 1516 destination ports to 69 destination services for the CPTC'17 dataset. Contextually, this has the effect of indicating what service is being targeted by attackers, rather than just knowing a specific port number. Herein the processed Destination Ports are referred to as Destination Services. Additionally, the dimensionality reduction step can easily be expanded or customized on a per network basis given a corporation's configuration of services.

Finally, a set of simple statistical criterion were used to segment timestamps into bins. Traditional modeling of cyber attacks use attack stages to segment actions into a series of contiguous stages with dependencies on previous stages. The beginning of an attack may consist of reconnaissance based actions, yielding information about which IP to target in later attack stages. Similarly, the CPTC dataset may be segmented to try and capture unique behaviors into different Time Bins.

Following the methodology shown by \cite{us} bins were generated by smoothing the histogram timestamps and taking the first derivative to identify local minima and maxima. Then stages were cut if they contained at least $10\%$ of the total data and consecutive events at the candidate point contained less than $0.5\%$ of the total data. The goal of this ruleset was to capture significantly different types of traffic that does not split bursts of data into multiple stages.

Table \ref{table:features} shows the number of unique values present for each target IP address tested after preprocessing the data for CPTC 2017. Additionally a single character symbol is defined for each feature in parenthesis to compact future plot labels. The same set of features and preprocessing steps were used for CPTC 2018.

\begin{table}[!htbp]
    \caption{Number of Unique Feature Values for Assorted Target IPs}
    \label{table:features}
    \centering
    \begin{tabular}{l|cccc}
    & \multicolumn{4}{c}{\textbf{Machine IP Address}}  \\
    \multicolumn{1}{l|}{} & \multicolumn{1}{l}{\textbf{10.0.0.100}} & \multicolumn{1}{l}{\textbf{10.0.0.27}} & \multicolumn{1}{l}{\textbf{10.0.0.22}} & \multicolumn{1}{l|}{\textbf{10.0.99.143}} \\ \hline
        \textbf{Alerts} & \multicolumn{1}{c|}{3388} & \multicolumn{1}{c|}{3186} & \multicolumn{1}{c|}{2974} & \multicolumn{1}{c|}{2182} \\
        \textbf{Alert Signatures (A)} & \multicolumn{1}{c|}{53} & \multicolumn{1}{c|}{41} & \multicolumn{1}{c|}{34} & \multicolumn{1}{c|}{38} \\ 
        \textbf{Destination Service (D)} & \multicolumn{1}{c|}{25} & \multicolumn{1}{c|}{27} & \multicolumn{1}{c|}{21} & \multicolumn{1}{c|}{23} \\  
        \textbf{Source IPs (S)} & \multicolumn{1}{c|}{6} & \multicolumn{1}{c|}{6} & \multicolumn{1}{c|}{6} & \multicolumn{1}{c|}{7} \\ 
        \textbf{Timebins (T)} & \multicolumn{1}{c|}{36} & \multicolumn{1}{c|}{28} & \multicolumn{1}{c|}{30} & \multicolumn{1}{c|}{31} \\ 
    \end{tabular}
\end{table}

It is worth noting that if applied to real world attack data, segmenting Source IP addresses by subnet could provide useful context to the originator of the attack. However, if an attack was well planned or socially engineered to look like normal behavior within that portion of the network then no alerts would be generated. These shortcomings remain outstanding challenges of NIDS as a whole and fall outside the scope of this work.  

\subsection{Histogram Intersection}
\label{sec:inter}

The Histogram Intersection metric compares the similarity of two histograms within the same domain by computing the amount of overlap between them. It is naturally bounded between 0 and 1, intuitive to understand, extends to joint distributions of features, and may be graphed to directly visualize results. Let \emph{P} represent the ground truth data histogram and \emph{Q} represent the generated data histogram, each with $N$ samples. The Histogram Intersection ($G$) is then defined as shown in (\ref{eq:intersection}).

\begin{equation}
    \label{eq:intersection}
    G(P,Q) = {{\sum_{i=0}^{N} min(P_i, Q_i)} \over {max(\sum_{i=0}^{N} P_i, \sum_{i=0}^{N} Q_i)}}
\end{equation}

Additionally, combinations of features were considered in both the ground truth and generated datasets. For example, the histogram of all possible combinations of values for Alert Signature and Destination Service is one class of combinations. The intersection of these combinations of features was also taken by representing feature combinations as joint distributions; Herein, the number of features included in each histogram is referred to as an $m$-tuple. The intersection scores of varying $m$-tuple histograms were then reviewed to judge the performance of the generator in recreating increasingly complex data.

\subsection{Dependencies within Alert Features}
\label{sec:conditional}
An additional property of the intersection of histograms is that it reveals intra-alert feature dependencies, which helps explain and assess the performance of GAN models. This is accomplished by reviewing the difference in histogram intersection as new features are introduced to the histogram. That is, if the intersection for feature $X$ is 0.8 and the intersection for 2-tuple histogram of features $X$ and $Y$ is 0.79 then $X$ is expected to predict $Y$ well; this relationship is not necessarily to be bidirectional.

To confirm that these scores correctly indicate feature dependencies the conditional entropy of $Y$ given $X$ can be computed directly. It is important to extend the previous $m$-tuple notation to include a conditional equivalent. This is defined as the $Y|X$-tuple, which defines a single feature $Y$ given a vector of features $X$.

The conditional entropy was first weighted such that a single value could be obtained to represent all possible input condition values for a fixed set of features. It was computed using (\ref{eq:wce}). In this equation the weight term ${{|w_i|} \over {|w|}}$ represents the number of times the conditional input values at index $i$ occurred in the dataset, divided by the total number of alerts. ${p_{i|j}}$ represents the probability of the output feature value at index $j$ occurring given the input feature values at index $i$. 

\begin{equation}
    \label{eq:wce}
    \widehat{H}_{Y|X_0, X_1, ..., X_m} = \sum_{i=0}^{N} \bigg({{|w_i|} \over {|w|}} * {\sum_{j=0}^{Z}\Big({p_{i|j} * \log ({1 \over{p_{i|j}}})}\Big)}\bigg)
\end{equation}

This score was then normalized in (\ref{eq:norm_wce}) by dividing the weighted entropy by the entropy maximizing distribution for a discrete dataset with finite support; the uniform distribution $\mathbb{U}$ with cardinality equivalent to the number of unique elements in the joint distribution considered. Using this metric, entropy values close to 0 indicated that a given input resulted in a particular output with near determinism. Entropy values close to 1 indicated that given a particular input condition all outputs were equally probable. 

\begin{equation}
    \label{eq:norm_wce}
    \overline{H}_{Y|X_0, X_1, ..., X_m} = {{\widehat{H}_{Y|X_0, X_1, ..., X_m}} \over{ H(\mathbb{U})}}
\end{equation}

Similarly, the joint entropy was computed and normalized for all m-tuples using (\ref{eq:je}) and dividing by a uniform distribution. This metric provides a baseline for analyzing the results of the weighted normalized conditional entropy by illustrating the randomness of the data if feature dependence is not considered. 

\begin{equation}
    \label{eq:je}
    {H}_{X_m} = -\sum_{x_m} p(x_0, x_1,...,x_m)  * \log \big({{p(x_0,x_1,...,x_m)}}\big)
\end{equation}

Most importantly, comparing the weighted normalized conditional entropy of the ground truth and generated data provided a numerical way to evaluate how well the GAN learned to mimic feature interactions seen in real data. Normalizing the entropies makes it possible to draw comparisons between different input condition features, different input condition feature vector sizes, and different target IP addresses. Similarly, normalizing the joint entropy of the varying m-tuple feature combinations allows for a universal method to compare the challenge of recreating increasingly complex data. The direct comparison of normalized joint entropy and normalized conditional entropy further illustrates the importance of capturing feature dependencies with GANs.

\subsection{Output Modes and Attack Stages}
\label{sec:outputs}

Finally, the purpose of adding in the mutual information constrained model (WGAN-GPMI) was to palliate mode dropping. In order to evaluate this, a two step analysis process was employed. First, the number of output modes dropped for all feature combinations was collected. Then, to provide a result with direct contextual meaning to cybersecurity, the generated alerts were mapped to attack stages to show that WGAN-GPMI is capable of synthesizing alerts pertaining to more attack stages than the WGAN-GP model is. 

We define the attack stages based off the type of actions taken, such as reconnaissance and data exfiltration. Alert attributes such as category or signature gives the inclination of attack type; however, the category is an arbitrary high-level description of the attack type that may not accurately represent the outcome of the action whereas the signature may be at too fine of granularity to depict the attack behavior. Thus, this work also assesses the synthetically generated alerts by mapping the alert signatures to one of the defined attack stages given in Table \ref{tab:attack_stages}, by manually examining the objective and outcome described in the signature description.   

\begin{table}[hbtp!]
\caption{Attack Stage Types}
\centering
\label{tab:attack_stages}
    \begin{tabular}{|l||l|}
    \hline
    \multicolumn{1}{|c||}{Attack Stage} & \multicolumn{1}{c|}{Description}                               \\ \hline \hline
    IP Scan                                & Scan to reveal IP addresses                                    \\ \hline
    Service Scan                           & Scan to reveal services active on a target                     \\ \hline
    Targeted Scan                          & A targeted and specific scan on a machine                      \\ \hline
    Social Engineering                     & Deceiving / manipulating individuals for malicious intent \\ \hline
    Surfing                                & Browsing publicly available information to research target     \\ \hline
    Specific Exploits                      & Using a specific vulnerability on a target                     \\ \hline
    Escalate Privledges                    & Gaining unauthorized privileges                                \\ \hline
    Zero Day                               & Conducting an action not recorded and/or observed before       \\ \hline
    Malware Injection                      & Delivering malware to target                                   \\ \hline
    Degrade Operations                     & Reduce or interrupt ``normal" functionality of a target        \\ \hline
    Data Exfiltration                      & Steal and extract sensitive information                        \\ \hline
    \end{tabular}
\end{table}

Using this mapping we can see if GANs capture latent behaviors within the dataset even when it fails to output specific alert signatures that occurred explicitly in the dataset. Additionally, the output domain coverage for each model is shown to compare the model's performance on fine grained generation to that of the attack action distribution. 

\begin{table*}[!htbp]
	\caption{Histogram Intersection for all Feature Combinations: CPTC'17}
	\label{tab:inter}
	\centering
% 	\begin{adjustbox}{angle=90}
	\begin{tabular}{l|c|c|c|c|c|c|c|c|c|}
		\multicolumn{1}{c|}{} & \multicolumn{9}{c|}{\textbf{Target Machine IP Address}} \\
		\multicolumn{1}{c|}{} & \multicolumn{4}{c|}{\textbf{WGAN-GP}} &  & \multicolumn{4}{c|}{\textbf{WGAN-GPMI}} \\
		\multicolumn{1}{c|}{\textbf{Features}} & \textbf{10.0.0.100} & \textbf{10.0.0.27} & \textbf{10.0.0.22} & \textbf{10.0.99.143} & \textbf{} & \textbf{10.0.0.100} & \textbf{10.0.0.27} & \textbf{10.0.0.22} & \textbf{10.0.99.143} \\ \hline
		\textbf{A} & $0.697 \pm 0.002$ & $0.658 \pm 0.007$ & $0.844 \pm 0.005$ & \textbf{0.858 $\pm$ 0.009} & & \textbf{0.890 $\pm$ 0.006} & \textbf{0.833 $\pm$ 0.005} & \textbf{0.847 $\pm$ 0.006} & 0.808 $\pm$ 0.009 \\
		\textbf{D} & $0.772 \pm 0.007$ & $0.660 \pm 0.006$ & $0.843 \pm 0.005$ & \textbf{0.905 $\pm$ 0.009} & & \textbf{0.899 $\pm$ 0.006} & \textbf{0.846 $\pm$ 0.005} & 0.823 $\pm$ 0.007 & 0.827 $\pm$ 0.009 \\
		\textbf{S} & $0.717 \pm 0.007$ & $0.867 \pm 0.009$ & \textbf{0.846 $\pm$ 0.008} & $0.843 \pm 0.009$ & & \textbf{0.906 $\pm$ 0.008} & 0.909 $\pm$ 0.005 & 0.755 $\pm$ 0.005 & 0.881 $\pm$ 0.010 \\
		\textbf{T} & $0.814 \pm 0.008$ & $0.760 \pm 0.007$ & $0.818 \pm 0.008$ & $0.723 \pm 0.009$ & & \textbf{0.892 $\pm$ 0.008} & \textbf{0.815 $\pm$ 0.007} & 0.844 $\pm$ 0.008 & \textbf{0.819 $\pm$ 0.009} \\ \hline
		\textbf{A,T} & $0.668 \pm 0.007$ & $0.630 \pm 0.007$ & $0.741 \pm 0.007$ & $0.630 \pm 0.008$ & & \textbf{0.774 $\pm$ 0.008} & \textbf{0.774 $\pm$ 0.007} & 0.754 $\pm$ 0.008 & \textbf{0.717 $\pm$ 0.010} \\
		\textbf{A,S} & $0.634 \pm 0.007$ & $0.590 \pm 0.007$ & \textbf{0.774 $\pm$ 0.004} & $0.757 \pm 0.008$ & & \textbf{0.791 $\pm$ 0.008} & \textbf{0.747 $\pm$ 0.007} & 0.718 $\pm$ 0.008 & 0.771 $\pm$ 0.010 \\
		\textbf{S,D} & $0.698 \pm 0.007$ & $0.598 \pm 0.007$ & \textbf{0.768 $\pm$ 0.004} & $0.779 \pm 0.009$ & & \textbf{0.829 $\pm$ 0.007} & \textbf{0.758 $\pm$ 0.005} & 0.715 $\pm$ 0.005 & 0.800 $\pm$ 0.010 \\
		\textbf{D,T} & $0.710 \pm 0.007$ & $0.631 \pm 0.006$ & $0.768 \pm 0.004$ & $0.659 \pm 0.009$ & & \textbf{0.790 $\pm$ 0.008} & \textbf{0.777 $\pm$ 0.007} & 0.736 $\pm$ 0.008 & \textbf{0.726 $\pm$ 0.009} \\
		\textbf{S,T} & $0.710 \pm 0.007$ & $0.702 \pm 0.008$ & $0.741 \pm 0.007$ & $0.698 \pm 0.009$ & & \textbf{0.778 $\pm$ 0.008} & \textbf{0.791 $\pm$ 0.006} & 0.701 $\pm$ 0.007 & \textbf{0.782 $\pm$ 0.010} \\
		\textbf{A,D} & $0.693 \pm 0.002$ & $0.637 \pm 0.006$ & $0.828 \pm 0.005$ & \textbf{0.830 $\pm$ 0.008} & & \textbf{0.825 $\pm$ 0.008} & \textbf{0.822 $\pm$ 0.006} & 0.820 $\pm$ 0.008 & 0.777 $\pm$ 0.010 \\ \hline
		\textbf{A,S,T} & $0.599 \pm 0.007$ & $0.558 \pm 0.007$ & $0.686 \pm 0.007$ & $0.580 \pm 0.008$ & & \textbf{0.655 $\pm$ 0.008} & \textbf{0.727 $\pm$ 0.006} & 0.683 $\pm$ 0.008 & \textbf{0.632 $\pm$ 0.010} \\
		\textbf{A,S,D} & $0.627 \pm 0.007$ & $0.573 \pm 0.007$ & \textbf{0.761 $\pm$ 0.004} & $0.734 \pm 0.009$ & & \textbf{0.733 $\pm$ 0.008} & \textbf{0.737 $\pm$ 0.006} & 0.697 $\pm$ 0.008 & 0.740 $\pm$ 0.010 \\
		\textbf{A,D,T} & $0.653 \pm 0.007$ & $0.615 \pm 0.006$ & $0.756 \pm 0.007$ & $0.612 \pm 0.008$ & & \textbf{0.715 $\pm$ 0.008} & \textbf{0.731 $\pm$ 0.006} & 0.731 $\pm$ 0.007 & \textbf{0.685 $\pm$ 0.010} \\
		\textbf{S,D,T} & $0.611 \pm 0.007$ & $0.569 \pm 0.007$ & \textbf{0.690 $\pm$ 0.007} & $0.597 \pm 0.008$ & & 0.652 $\pm$ 0.008 & \textbf{0.734 $\pm$ 0.007} & 0.632 $\pm$ 0.008 & 0.635 $\pm$ 0.010 \\ \hline
		\textbf{A,S,D,T} & $0.584 \pm 0.007$ & $0.548 \pm 0.007$ & $0.601 \pm 0.007$ & $0.571 \pm 0.008$ & & \textbf{0.607 $\pm$ 0.008} & \textbf{0.718 $\pm$ 0.006} & 0.626 $\pm$ 0.008 & 0.615 $\pm$ 0.010 \\
	\end{tabular}
% 	\end{adjustbox}
\end{table*}

% Correct the standard deviations
\begin{table*}[!htbp]
	\caption{Histogram Intersection for all Feature Combinations: CPTC'18}
	\label{tab:inter18}
	\centering
% 	\begin{adjustbox}{angle=90}
		\begin{tabular}{l|c|c|c|c|c|c|c|c|c|}
			\multicolumn{1}{c|}{\textbf{}} & \multicolumn{9}{c|}{\textbf{Target Machine IP Address}} \\
			\multicolumn{1}{c|}{} & \multicolumn{4}{c|}{\textbf{WGAN-GP}} &  & \multicolumn{4}{c|}{\textbf{WGAN-GPMI}} \\
			\multicolumn{1}{c|}{\textbf{Features}} & \textbf{10.0.1.46} & \textbf{10.0.1.5} & \textbf{10.0.0.24} & \textbf{10.0.0.22} &  & \textbf{10.0.1.46} & \textbf{10.0.1.5} & \textbf{10.0.0.24} & \textbf{10.0.0.22} \\ \hline
			\textbf{A} & 0.752 $\pm$ 0.005 & \textbf{0.810} $\pm$ \textbf{0.003} & 0.801 $\pm$ 0.004 & 0.815 $\pm$ 0.004 &  & \textbf{0.852} $\pm$ \textbf{0.006} & 0.765 $\pm$ 0.005 & 0.825 $\pm$ 0.004 & 0.863 $\pm$ 0.003 \\
			\textbf{D} & 0.764 $\pm$ 0.004 & 0.954 $\pm$ 0.005 & 0.800 $\pm$ 0.006 & 0.820 $\pm$ 0.006 &  & \textbf{0.859} $\pm$ \textbf{0.005} & 0.909 $\pm$ 0.004 & \textbf{0.918} $\pm$ \textbf{0.003} & \textbf{0.874} $\pm$ \textbf{0.005} \\
			\textbf{S} & 0.744 $\pm$ 0.005 & 0.740 $\pm$ 0.006 & 0.789 $\pm$ 0.005 & 0.812 $\pm$ 0.006 &  & \textbf{0.844} $\pm$ \textbf{0.006} & 0.785 $\pm$ 0.004 & \textbf{0.872} $\pm$ \textbf{0.005} & \textbf{0.867} $\pm$ \textbf{0.003} \\
			\textbf{T} & 0.782 $\pm$ 0.007 & 0.718 $\pm$ 0.005 & 0.851 $\pm$ 0.006 & 0.811 $\pm$ 0.007 &  & 0.826 $\pm$ 0.005 & 0.766 $\pm$ 0.006 & \textbf{0.928} $\pm$ \textbf{0.004} & 0.857 $\pm$ 0.005 \\ \hline
			\textbf{A,T} & 0.764 $\pm$ 0.006 & 0.954 $\pm$ 0.004 & 0.800 $\pm$ 0.006 & \textbf{0.820} $\pm$ \textbf{0.005} &  & \textbf{0.859} $\pm$ \textbf{0.006} & 0.909 $\pm$ 0.004 & \textbf{0.918} $\pm$ \textbf{0.007} & 0.874 $\pm$ 0.004 \\
			\textbf{A,S} & 0.744 $\pm$ 0.005 & 0.740 $\pm$ 0.006 & 0.789 $\pm$ 0.007 & 0.812 $\pm$ 0.006 &  & \textbf{0.844} $\pm$ \textbf{0.004} & 0.785 $\pm$ 0.005 & \textbf{0.872} $\pm$ \textbf{0.004} & \textbf{0.867} $\pm$ \textbf{0.006} \\
			\textbf{S,D} & 0.782 $\pm$ 0.006 & 0.718 $\pm$ 0.006 & 0.851 $\pm$ 0.005 & 0.811 $\pm$ 0.004 &  & \textbf{0.826} $\pm$ \textbf{0.005} & 0.766 $\pm$ 0.006 & \textbf{0.928} $\pm$ \textbf{0.005} & 0.857 $\pm$ 0.006 \\
			\textbf{D,T} & 0.679 $\pm$ 0.004 & 0.701 $\pm$ 0.005 & 0.784 $\pm$ 0.005 & 0.793 $\pm$ 0.006 &  & \textbf{0.746} $\pm$ \textbf{0.005} & 0.744 $\pm$ 0.006 & \textbf{0.898} $\pm$ \textbf{0.007} & 0.811 $\pm$ 0.006 \\
			\textbf{S,T} & 0.667 $\pm$ 0.005 & 0.734 $\pm$ 0.006 & 0.766 $\pm$ 0.007 & 0.789 $\pm$ 0.005 &  & \textbf{0.753} $\pm$ \textbf{0.006} & 0.773 $\pm$ 0.006 & \textbf{0.848} $\pm$ \textbf{0.007} & 0.816 $\pm$ 0.005 \\
			\textbf{A,D} & 0.646 $\pm$ 0.007 & 0.644 $\pm$ 0.007 & 0.774 $\pm$ 0.006 & 0.796 $\pm$ 0.007 &  & 0.694 $\pm$ 0.006 & 0.646 $\pm$ 0.005 & \textbf{0.862} $\pm$ \textbf{0.006} & 0.807 $\pm$ 0.004 \\ \hline
			\textbf{A,S,T} & 0.679 $\pm$ 0.006 & 0.701 $\pm$ 0.004 & 0.784 $\pm$ 0.005 & 0.793 $\pm$ 0.006 &  & \textbf{0.746} $\pm$ \textbf{0.006} & 0.744 $\pm$ 0.005 & \textbf{0.898} $\pm$\textbf{0.005} & 0.811 $\pm$ 0.006 \\
			\textbf{A,S,D} & 0.667 $\pm$ 0.005 & 0.734 $\pm$ 0.004 & 0.766 $\pm$ 0.003 & 0.789 $\pm$ 0.004 &  & \textbf{0.753} $\pm$ \textbf{0.005} & 0.773 $\pm$ 0.006 & \textbf{0.848} $\pm$ \textbf{0.006} & 0.816 $\pm$ 0.005 \\
			\textbf{A,D,T} & 0.646 $\pm$ 0.007 & 0.644 $\pm$ 0.007 & 0.774 $\pm$ 0.005 & 0.796 $\pm$ 0.006 &  & 0.694 $\pm$ 0.006 & 0.646 $\pm$ 0.007 & \textbf{0.862} $\pm$ \textbf{0.005} & 0.807 $\pm$ 0.005 \\
			\textbf{S,D,T} & 0.536 $\pm$ 0.006 & 0.617 $\pm$ 0.006 & 0.750 $\pm$ 0.005 & \textbf{0.769} $\pm$ \textbf{0.005} &  & 0.580 $\pm$ 0.006 & 0.616 $\pm$ 0.005 & \textbf{0.820} $\pm$ \textbf{0.005} & 0.762 $\pm$ 0.005 \\ \hline
			\textbf{A,S,D,T} & 0.536 $\pm$ 0.006 & 0.617 $\pm$ 0.005 & 0.750 $\pm$ 0.004 & \textbf{0.769} $\pm$ \textbf{0.005} &  & 0.580 $\pm$ 0.006 & 0.616 $\pm$ 0.007 & \textbf{0.820} $\pm$ \textbf{0.006} & 0.762$\pm$ 0.006 
		\end{tabular}
% 	\end{adjustbox}
\end{table*}

\section{Results and Analysis}
\label{sec:results}

Each the models used were trained using slightly different hyper parameter settings. WGAN-GP was trained for a total of $200$ epochs, while the WGAN-GPMI model was trained for $300$ epochs. This increase in the number of epochs used was to account for the added complexity of training the Estimator network in conjunction with the Discriminator and Generator. Additionally, the lambda value controlling the gradient penalty term was set to $0.1$ and $0.4$ for the WGAN-GP and WGAN-GPMI models respectively. All other hyperparameter values were held constant for both models. The ADAM optimizer was used with $learning\_rate = 5e-5$, $\beta_1=0.5$ and $\beta_2=0.8$ for batches of $100$ alerts. 

\subsection{Histogram Intersection}
\label{sec:histo_res}

The accuracy of each model was observed using the intersection of histograms metric discussed in Section \ref{sec:inter}. The Histogram Intersections ($G$-scores) were computed for several targets when using both WGAN-GP and WGAN-GPMI. The maximum intersection score for each combination of features was bolded if the given score was at least 0.05 greater than the intersection score of the corresponding result from the other model. Each model was sampled $1000$ times to compute the standard deviation of the $G$-scores. Table \ref{tab:inter} and Table \ref{tab:inter18} show the G-scores and standard deviation for CPTC'17 and CPTC'18 respectively. 

First, note that both WGAN-GP and WGAN-GPMI achieves reasonably good performance, even when considering the combination of all 4 feature values; Samples from WGAN-GP are able to achieve up to 60\% intersection with the ground truth distribution while samples generated by the WGAN-GPMI model achieve up to 71\% intersection in CPTC'17; similarly the CPTC'18 data shows improvements when using the WGAN-GPMI model and a maximum $G$-score of 82\%. Secondly, note that for both IP addresses the Mutual Information constraint in the WGAN-GPMI model is able to increase the $G$-score. This is a result of the model learning a probability distribution which is closer to that of the ground truth data. 

It is interesting to note that the effect of the mutual information constraint varies from target IP address to target IP address. For example, target 10.0.0.22 only has small improvements to Histogram Intersection when using the WGAN-GPMI model. In several cases, such as Source IP and Destination Service, the intersection score actually drops. On the other hand Target IP addresses such as 10.0.0.27 see a large benefit from using the mutual information constraint. On average, the Histogram Intersection is $14.63$\% higher for the WGAN-GPMI model than it's WGAN-GP counterpart when trained on the CPTC'17 dataset. When trained on CPTC'18 the average increase in $G$-score is 5.6\%.  These results are particularly interesting, as the intent of the mutual information constraint is to improve mode dropping in the generator, not to directly improve Histogram Intersection. Palliating mode dropping is indirectly related to increasing Histogram Intersection in many cases because it distributes output sample entropy across more output values than standard GAN models do when exhibiting mode dropping. 

Another interesting result of Table \ref{tab:inter} is that the intersection of histograms is resilient to earlier score bias. Consider the intersection score of Timestamp (T) on target IP 10.0.0.22. This feature has a high $G$-score of 81.8\%, potentially leading to the fallacious expectation that any combination with T will also score high. When moving to testing 2-tuple combinations such as Timestamp (T) + Source IP (S) however the intersection drops significantly. This is due to a lack of dependence between T and S. 

Finally, it is important to note that the Histogram Intersections are generally higher for the data synthesized to emulate CPTC'18 than the data emulating CPTC'17. One cause of this is the increased number of alerts available for each of the targets in CPTC'18 however further study is needed to determine the exact effects of data set size on model output. Additionally, these results demonstrate the promise of applying these network architectures to various cyber alert datasets, not just the one's presented here.

\subsection{Dependencies within Alert Features}

\begin{figure}[!htbp]
    \centering
    \includegraphics[width=\columnwidth]{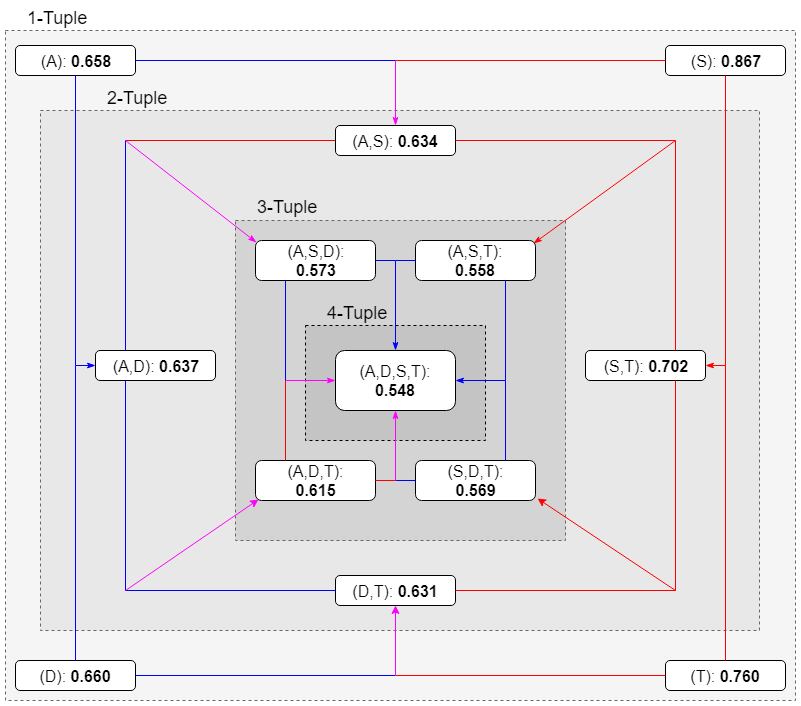}
    \caption{Intersection of Histogram Scores for combinations of feature inputs on target IP 10.0.0.22. \vspace*{-10pt}}
    \label{fig:intersect_graph}
\end{figure}

Given the intersection scores in Tables \ref{tab:inter} and \ref{tab:inter18} it is possible to construct a graph of intersection scores as varying values of $m$ are used for the $m$-tuple histograms. By representing the data is this manner it becomes easier to visualize feature dependencies based off the difference between the $m$-tuple intersection scores and $m+1$-tuple intersection scores. Such a graph was constructed for target IP 10.0.0.22, using the WGAN-GP model results, in Fig. \ref{fig:intersect_graph}.

\begin{figure}[!htbp]
    \centering
    \includegraphics[width=\columnwidth]{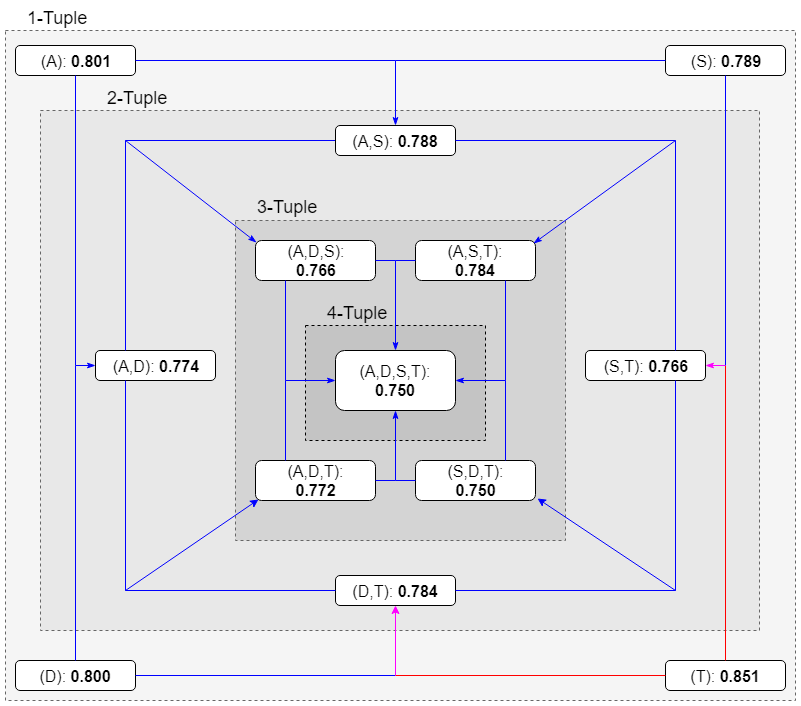}
    \caption{Intersection of Histogram Scores for combinations of feature inputs on target IP 10.0.0.24. \vspace*{-10pt}}
    \label{fig:404}
\end{figure}

\begin{table*}[!htbp]
	\caption{Weighted Normalized Conditional Entropy Values for all target IPs: WGAN-GPMI Result}
	\label{tab:ce}
	\centering
		\begin{tabular}{l|c|c|c|c|c|c|c|c|c|}
			\multicolumn{1}{c|}{} & \multicolumn{9}{c|}{\textbf{Target Machine IP Address}} \\
			\multicolumn{1}{c|}{} & \multicolumn{4}{c|}{\textbf{Ground Truth Results}} &  & \multicolumn{4}{c|}{\textbf{Generated Results}} \\
			\multicolumn{1}{c|}{\textbf{Features}} & \textbf{10.0.0.100} & \textbf{10.0.0.27} & \textbf{10.0.0.22} & \textbf{10.0.99.143} &  & \textbf{10.0.0.100} & \textbf{10.0.0.27} & \textbf{10.0.0.22} & \textbf{10.0.99.143} \\ \cline{1-10}
			\multicolumn{1}{l|}{\textbf{A{\given}T}} & \textbf{0.244} & \textbf{0.238} & \textbf{0.153} & \textbf{0.333} &  & \textbf{0.244} & \textbf{0.238} & \textbf{0.153} & \textbf{0.334} \\
			\multicolumn{1}{l|}{\textbf{T{\given}S}} & \textbf{0.593} & \textbf{0.463} & \textbf{0.515} & \textbf{0.695} &  & \textbf{0.593} & \textbf{0.463} & \textbf{0.516} & \textbf{0.695} \\
			\multicolumn{1}{l|}{\textbf{T{\given}A}} & \textbf{0.330} & \textbf{0.339} & \textbf{0.695} & \textbf{0.246} &  & \textbf{0.330} & \textbf{0.339} & \textbf{0.695} & \textbf{0.246} \\
			\multicolumn{1}{l|}{\textbf{S{\given}T}} & \textbf{0.262} & 0.252 & \textbf{0.263} & \textbf{0.405} &  & \textbf{0.263}	& 0.186	& \textbf{0.252}	& \textbf{0.406} \\
			\multicolumn{1}{l|}{\textbf{S{\given}A}} & 0.800 & 0.752 & 0.831 & 0.711 &  & 0.229	& 0.239	& 0.526	& 0.222 \\
			\multicolumn{1}{l|}{\textbf{D{\given}S}} & 0.346 & 0.445 & 0.253 & 0.278 &  & 0.509	& 0.385	& 0.207	& 0.558 \\
			\multicolumn{1}{l|}{\textbf{A{\given}D}} & 0.080 & 0.222 & 0.070 & 0.288 &  & 0.149	& 0.026	& 0.007	& 0.097 \\
			\multicolumn{1}{l|}{\textbf{T{\given}D}} & \textbf{0.479} & \textbf{0.346} & \textbf{0.655} & \textbf{0.383} &  & \textbf{0.479} & \textbf{0.346} & \textbf{0.655} & \textbf{0.383} \\
			\multicolumn{1}{l|}{\textbf{D{\given}T}} & \textbf{0.287} & \textbf{0.234} & \textbf{0.152} & \textbf{0.379} &  & \textbf{0.287} & \textbf{0.234} & \textbf{0.152} & \textbf{0.379} \\
			\multicolumn{1}{l|}{\textbf{A{\given}S}} & 0.346 & \textbf{0.385} & 0.271 & 0.475 &  & 0.403	& \textbf{0.376}	& 0.214	& 0.543 \\
			\multicolumn{1}{l|}{\textbf{S{\given}D}} & 0.822 & 0.779 & 0.856 & 0.785 &  & 0.436	& 0.260	& 0.474	& 0.301 \\
			\multicolumn{1}{l|}{\textbf{D{\given}A}} & 0.006 & 0.246 & 0.006 & 0.016 &  & 0.055	& 0.009	& 0.048	& 0.000 \\ \hline
			\multicolumn{1}{l|}{\textbf{S{\given}A,D}} & 0.799 & 0.747 & 0.829 & 0.705 &  & 0.228 &	0.226 &	0.474 &	0.222 \\
			\multicolumn{1}{l|}{\textbf{D{\given}S,T}} & \textbf{0.171} & \textbf{0.118} & \textbf{0.013} & \textbf{0.149} &  & \textbf{0.171} & \textbf{0.118} & \textbf{0.013} & \textbf{0.149} \\
			\multicolumn{1}{l|}{\textbf{S{\given}D,T}} & \textbf{0.107} & \textbf{0.025} & \textbf{0.101} & \textbf{0.024} &  & \textbf{0.107} & \textbf{0.025} & \textbf{0.101} & \textbf{0.024} \\
			\multicolumn{1}{l|}{\textbf{T{\given}A,D}} & \textbf{0.316} & \textbf{0.335} & \textbf{0.650} & \textbf{0.246} &  & \textbf{0.316} & \textbf{0.335} & \textbf{0.650} & \textbf{0.246} \\
			\multicolumn{1}{l|}{\textbf{A{\given}S,D}} & 0.069 & 0.206 & 0.056 & 0.245 &  & 0.018 &	0.003 &	0.007 &	0.055 \\
			\multicolumn{1}{l|}{\textbf{A{\given}S,T}} & 0.131 & \textbf{0.117} & \textbf{0.013} & \textbf{0.130} &  & 0.112 &	\textbf{0.117} &	\textbf{0.013} &	\textbf{0.131} \\
			\multicolumn{1}{l|}{\textbf{A{\given}D,T}} & \textbf{0.038} & \textbf{0.018} & \textbf{0.001} & \textbf{0.004} &  & \textbf{0.038} & \textbf{0.018} & \textbf{0.001} & \textbf{0.004} \\
			\multicolumn{1}{l|}{\textbf{D{\given}A,S}} & 0.005 & 0.243 & 0.005 & 0.012 &  & 0.054 &	0.000 &	0.000 &	0.000 \\
			\multicolumn{1}{l|}{\textbf{T{\given}S,D}} & 0.393 & 0.340 & 0.587 & 0.348 &  & 0.176 &	0.144 &	0.334 &	0.170 \\
			\multicolumn{1}{l|}{\textbf{D{\given}A,T}} & 0.004 & 0.238 & 0.003 & 0.007 &  & 0.044 &	0.006 &	0.000 &	0.000 \\
			\multicolumn{1}{l|}{\textbf{S{\given}A,T}} & 0.211 & 0.178 & \textbf{0.100} & 0.228 &  & 0.055 &	0.012 &	\textbf{0.100} &	0.019 \\
			\multicolumn{1}{l|}{\textbf{T{\given}A,S}} & 0.365 & 0.312 & 0.561 & 0.302 &  & 0.170 &	0.144 &	0.328 &	0.089 \\ \hline
			\multicolumn{1}{l|}{\textbf{A{\given}S,D,T}} & 0.041 & 0.172 & 0.028 & 0.195 &  & 0.005	& 0.003	& 0.000	& 0.001 \\
			\multicolumn{1}{l|}{\textbf{D{\given}A,S,T}} & 0.002 & 0.232 & 0.002 & 0.004 &  & 0.044	& 0.000	& 0.000	& 0.000 \\
			\multicolumn{1}{l|}{\textbf{T{\given}A,D,S}} & 0.209 & 0.167 & 0.498 & 0.222 &  & 0.157	& 0.144	& 0.328	& 0.089 \\
			\multicolumn{1}{l|}{\textbf{S{\given}A,T,D}} & 0.362 & 0.302 & 0.558 & 0.294 &  & 0.055	& 0.004	& 0.100	& 0.019
		\end{tabular}
\end{table*}

\begin{table*}[!htbp]
	\caption{Normalized Joint Entropy Values for all target IPs: WGAN-GPMI Result}
	\label{tab:je}
	\centering
% 	\begin{adjustbox}{angle=90}
		\begin{tabular}{l|c|c|c|c|c|c|c|c|c|}
			\multicolumn{1}{c|}{} & \multicolumn{9}{c|}{\textbf{Target Machine IP Address}} \\
			\multicolumn{1}{c|}{} & \multicolumn{4}{c|}{Ground Truth Results} &  & \multicolumn{4}{c|}{Generated Results} \\
			\multicolumn{1}{c|}{Features} & \textbf{10.0.0.100} & \textbf{10.0.0.27} & \textbf{10.0.0.22} & \textbf{10.0.99.143} &  & \textbf{10.0.0.100} & \textbf{10.0.0.27} & \textbf{10.0.0.22} & \textbf{10.0.99.143} \\ \hline
			\textbf{A,D} & 0.715 & 0.652 & 0.469 & 0.747 &  & 0.492 & 0.398 & 0.191 & 0.553 \\
			\textbf{A,S} & 0.704 & 0.683 & 0.674 & 0.749 &  & 0.462 & 0.531 & 0.456 & 0.652 \\
			\textbf{A,T} & 0.783 & 0.781 & 0.839 & 0.750 &  & 0.617 & 0.603 & 0.648 & 0.638 \\
			\textbf{D,T} & 0.811 & 0.807 & 0.858 & 0.767 &  & 0.682 & 0.592 & 0.608 & 0.672 \\
			\textbf{S,D} & 0.764 & 0.711 & 0.686 & 0.738 &  & 0.553 & 0.533 & 0.444 & 0.652 \\
			\textbf{S,T} & 0.816 & 0.801 & 0.842 & 0.773 &  & 0.672 & 0.642 & 0.665 & 0.593 \\ \hline
			\textbf{A,D,T} & 0.805 & 0.775 & 0.839 & 0.750 &  & 0.603 & 0.538 & 0.547 & 0.595 \\
			\textbf{A,S,D} & 0.735 & 0.683 & 0.674 & 0.749 &  & 0.508 & 0.478 & 0.389 & 0.579 \\
			\textbf{A,S,T} & 0.780 & 0.750 & 0.810 & 0.740 &  & 0.595 & 0.595 & 0.595 & 0.620 \\
			\textbf{S,D,T} & 0.806 & 0.775 & 0.810 & 0.748 &  & 0.646 & 0.595 & 0.581 & 0.647 \\ \hline
			\textbf{A,S,D,T} & 0.803 & 0.750 & 0.810 & 0.740 &  & 0.605 & 0.563 & 0.553 & 0.611
			\end{tabular}
% 	\end{adjustbox}
\end{table*}

Beginning from the outer edges of the graph, single-feature histograms are considered. Then each non-labeled vertex represents a union operator and connects to an $m+1$ feature joint distribution node closer to the center of the graph. This process continues until the center of the graph is reached. At the central node all four features are considered in the histogram computation. Lines are color coded such that blue lines indicate feature unions which result in less than 5\% difference between histogram scores. Conversely, lines which are red indicate an difference that is greater than 5\%. Recall that the standard deviation for individual scores was never higher than 0.01\%, meaning that the 5\% cutoffs here are notably outside the margin of noise. Lines which are purple indicate that only one of the original m-tuples has a drop in intersection greater than 5\%. All dashed lines are bounding boxes added to clearly segment the varying $m$-tuple histograms. 

Note that in order to maintain planarity of the graph, two of the 2-tuple intersections were removed in each graph. This was done to increase readability of the graph for the aforementioned explanation; the results for all combinations are still important to consider for the subsequent analysis.

To verify the aforementioned feature dependencies the weighted normalized conditional entropy is computed for all target IPs from the CPTC'17 dataset, across all potential $Y|X$ feature-tuples. These results are shown in Table \ref{tab:ce}. Additionally, all normalized joint entropy values are computed for all target IPs in \ref{tab:je} to provide a baseline representation of the amount of randomness in the data.

By computing the conditional entropy in Table \ref{tab:ce} it is apparent that the WGAN-GPMI model closely imitates the dependencies of the ground truth. In fact, several of the small valued m-tuples such as A{\given}T, T{\given}D, and D{\given}S,T all have identical conditional entropy values to the ground truth distribution. These values, as well as those within within 10\% of the ground truth entropy value, are highlighted for clarity.  

Using the graph presented in Fig. \ref{fig:intersect_graph} along with Table \ref{tab:ce} and Table \ref{tab:je}, several 3-tuple feature dependencies are reviewed for target IP 10.0.0.22. The joint features of interest include $A+D+T$ and $A+D+S$ and all potential conditional inputs utilizing these features.

Beginning with the joint of $A,D,S$, the graph from Fig. \ref{fig:intersect_graph} highlights two conditional entropy values of interest; $S|A,D$ and $D|A,S$. Observing the graph and difference in histogram intersection scores, $D$ should be easily predicted given $A$ and $S$. The conditional entropy for $D|A,S$ is low, as well as significantly lower than the joint entropy for $A,D,S$, matching expectation. Similarly $S|A,D$ matches expectations as the conditional entropy is much higher along with the difference in histogram intersection scores.

The joint of $A,D,T$ is less clear to analyze given that the conditional entropy of $T|A,D$ is high while the difference of intersection scores is low. One potential explanation for this comes from Table \ref{tab:je}. The joint distribution of $A,D$ has low entropy in both the ground truth and generated datasets. Then, the addition of feature $T$ drastically raises the entropy for both dataset. Given that both the conditional entropy and joint entropy are high it may be concluded that the distribution of $T$ is both stochastic and independent of the joint distribution $A,D$. Given this information the generator could learn to output values of $T$ randomly, maintaining a high intersection score score despite a lack of dependence with other alert features.

These graphs were also generated to analyze dependencies in the CPTC'18 data. Fig. \ref{fig:404} shows the results for target IP 10.0.0.24. Note that the feature dependencies for this data are considerably stronger than in the example from CPTC'17. The intersection scores have a much lower amount of variability between each level of m-tuples, indicating a higher level of intra-alert feature dependencies. In fact the delta from the highest $G$-score in the graph to the lowest is only an 11.1\% decrease. Three out of the four target IPs tested from CPTC'18 demonstrated small overall variability.

\subsection{Output Modes and Attack Stages}
\label{sec:as_res}

\begin{table*}[!htbp]
	\centering
	\caption{Output Modes Dropped and Noisy Outputs: CPTC'17}
	\label{tab:output_modes}
	\begin{tabular}{l|c|c|c|c|c|c|c|c|c|}
		\multicolumn{1}{c|}{} & \multicolumn{9}{c|}{\textbf{Target Machine IP Address}} \\
		\multicolumn{1}{c|}{} & \multicolumn{4}{c|}{\textbf{WGAN-GP}} &  & \multicolumn{4}{c|}{\textbf{WGAN-GPMI}} \\
		\multicolumn{1}{c|}{\textbf{Features}} & \textbf{10.0.0.100} & \textbf{10.0.0.27} & \textbf{10.0.0.22} & \textbf{10.0.99.143} & \textbf{} & \textbf{10.0.0.100} & \textbf{10.0.0.27} & \textbf{10.0.0.22} & \textbf{10.0.99.143} \\ \hline
	    \multicolumn{1}{c|}{\textbf{Noise}} & \multicolumn{1}{c|}{168} & \multicolumn{1}{c|}{235} & \multicolumn{1}{c|}{76} & \multicolumn{1}{c|}{321} & & \multicolumn{1}{c|}{213} & \multicolumn{1}{c|}{97} & \multicolumn{1}{c|}{107} & 98 \\
		\hline
		\multicolumn{1}{c|}{\textbf{Dropped}} & \multicolumn{1}{c|}{21} & \multicolumn{1}{c|}{15} & \multicolumn{1}{c|}{9} & \multicolumn{1}{c|}{10} & & \multicolumn{1}{c|}{10} & \multicolumn{1}{c|}{14} & \multicolumn{1}{c|}{4} & 8 \\ \hline
		\hline
		
		\multicolumn{1}{l|}{\textbf{\# Alerts}} & \multicolumn{1}{c|}{3388} & \multicolumn{1}{c|}{3166} & \multicolumn{1}{c|}{2974} & \multicolumn{1}{c|}{2182} & & \multicolumn{1}{l|}{} & \multicolumn{1}{l|}{} & \multicolumn{1}{l|}{} & \multicolumn{1}{l|}{} \\
		\multicolumn{1}{l|}{\textbf{\# Unique Modes}} & \multicolumn{1}{c|}{32} & \multicolumn{1}{c|}{27} & \multicolumn{1}{c|}{22} & \multicolumn{1}{c|}{27} & & \multicolumn{1}{l|}{} & \multicolumn{1}{l|}{} & \multicolumn{1}{l|}{} & \multicolumn{1}{l|}{} \\
		\multicolumn{1}{l|}{\textbf{\% Modes Dropped}} & \multicolumn{1}{c|}{0.6563} & \multicolumn{1}{c|}{0.5556} & \multicolumn{1}{c|}{0.4091} & 
		\multicolumn{1}{c|}{0.3704} & &  \multicolumn{1}{c|}{0.3125} & \multicolumn{1}{c|}{0.5185} & \multicolumn{1}{c|}{0.1818} & \multicolumn{1}{c|}{0.2963}\\
		\multicolumn{1}{l|}{\textbf{Noise Ratio}} & 5.250 & 8.704 & 3.455 & 11.889 & & 6.656 & 3.593 & 4.864 & 3.630
	\end{tabular}
\end{table*}

\begin{table*}[!htbp]
	\centering
	\caption{Output Modes Dropped and Noisy Outputs: CPTC'18}
	\label{tab:output_modes18}
	\begin{tabular}{l|c|c|c|c|c|c|c|c|c|}
	\multicolumn{1}{c|}{\textbf{}} & \multicolumn{9}{c|}{\textbf{Target Machine IP Address}} \\
			\multicolumn{1}{c|}{} & \multicolumn{4}{c|}{\textbf{WGAN-GP}} &  & \multicolumn{4}{c|}{\textbf{WGAN-GPMI}} \\
			\multicolumn{1}{c|}{\textbf{Features}} & \textbf{10.0.1.46} & \textbf{10.0.1.5} & \textbf{10.0.0.24} & \textbf{10.0.0.22} &  & \textbf{10.0.1.46} & \textbf{10.0.1.5} & \textbf{10.0.0.24} & \textbf{10.0.0.22} \\ \hline
	    \multicolumn{1}{c|}{\textbf{Noise}} & \multicolumn{1}{c|}{138} & \multicolumn{1}{c|}{59} & \multicolumn{1}{c|}{69} & \multicolumn{1}{c|}{69} & & \multicolumn{1}{c|}{92} & \multicolumn{1}{c|}{37} & \multicolumn{1}{c|}{87} & \multicolumn{1}{c|}{34} \\
		\hline
		\multicolumn{1}{c|}{\textbf{Dropped}} & \multicolumn{1}{c|}{7} & \multicolumn{1}{c|}{14} & \multicolumn{1}{c|}{10} & \multicolumn{1}{c|}{8} & & \multicolumn{1}{c|}{29} & \multicolumn{1}{c|}{12} & \multicolumn{1}{c|}{18} & \multicolumn{1}{c|}{18} \\ \hline
		\hline
		\multicolumn{1}{l|}{\textbf{\# Alerts}} & \multicolumn{1}{c|}{7475} & \multicolumn{1}{c|}{8695} & \multicolumn{1}{c|}{9861} & \multicolumn{1}{c|}{7996} & & \multicolumn{1}{l|}{} & \multicolumn{1}{l|}{} & \multicolumn{1}{l|}{} & \multicolumn{1}{l|}{} \\
		\multicolumn{1}{l|}{\textbf{\# Unique Modes}} & \multicolumn{1}{c|}{33} & \multicolumn{1}{c|}{31} & \multicolumn{1}{c|}{22} & \multicolumn{1}{c|}{29} & & \multicolumn{1}{l|}{} & \multicolumn{1}{l|}{} & \multicolumn{1}{l|}{} & \multicolumn{1}{l|}{} \\
		\multicolumn{1}{l|}{\textbf{\% Modes Dropped}} & \multicolumn{1}{c|}{0.2121} & \multicolumn{1}{c|}{0.4516} & \multicolumn{1}{c|}{0.4545} & 
		\multicolumn{1}{c|}{0.2759} & & \multicolumn{1}{c|}{0.8788} & \multicolumn{1}{c|}{0.3871} & \multicolumn{1}{c|}{0.8182} & \multicolumn{1}{c|}{0.6201}\\
		\multicolumn{1}{l|}{\textbf{Noise Ratio}} & 4.182 & 1.903 & 3.136 & 2.379 & & 2.788 & 1.194 & 3.955 & 1.172
	\end{tabular}
\end{table*}

Finally, to assess output modes captured by each model, we examine the number of output modes dropped or added by WGAN-GPMI. This was done by looking at all the unique alert feature combinations across A/D/S/T values that existed in the ground truth dataset versus those existed in the generated dataset. These sets of unique values were compared to see which modes were dropped, which were covered, and which existed in the generated set but not the ground truth set. We refer to these values as \emph{Dropped}, \emph{Covered}, and \emph{Noisy} respectively. 

Table \ref{tab:output_modes} shows the number of Dropped and Noisy outputs for each GAN model when trained on CPTC'17 data. The bottom four rows show the number of alerts, the number of unique 4-feature combinations, \% of output modes dropped, and ratio of noisy outputs to outputs within the domain of the ground truth.  Note that this table shows the direct benefit of mutual information maximization, as the number of output modes missed by the model decreases across the board for the WGAN-GPMI model. Some of the target IP addresses learn more output modes than others when moving to the WGAN-GPMI model; 10.0.0.100, as well as 10.0.0.22, halve the number of output modes dropped. On the other hand, 10.0.0.27 and 10.0.99.143 only see a minor improvement when adding in the mutual information constraint. These IP addresses instead see a large decrease in the number of noisy output modes when used for training the WGAN-GPMI model instead of the WGAN-GP model. It's also important to note that these output modes aren't inherently wrong since the individual feature value do exist in the ground truth dataset. However, there should be no gradient feedback to encourage the generation of these combinations of feature values since they don't occur in the ground truth dataset.

Table \ref{tab:output_modes18} shows the equivalent information for each generative model trained on the CPTC'18 data. Interestingly for all but target IP 10.0.0.24 the inverse relationship between Dropped and Noisy output modes holds true. Overall, the WGAN-GPMI model contributes to significantly lower amounts of noisy outputs when trained on the CPTC'18 data rather than palliating mode dropping. Despite this, Table \ref{tab:inter18} showed an increase in the intersection between the ground truth and generated histograms for the WGAN-GPMI model. This points to the decreased noisy alerts of this model resulting in a probability distribution that is significantly closer to the ground truth distribution despite missing more output modes.

One potential explanation for this would be that the missing output modes occur with such low probability that the loss score benefits when the model learns to overemphasize outputs which occur frequently in the ground truth. Methods to recreate rare samples remains an ongoing challenge that is incredibly important to the field of Cyber Security. It is possible that critical actions, such as data exfiltration may only generate a small number of alerts rarely over the course of an attack. Being able to model such behaviors would be extremely beneficial to proactive Cyber Defense, where critical vulnerabilities are identified and patched. 

\begin{figure*}[!htbp]
    \centering
    \includegraphics[width=\textwidth]{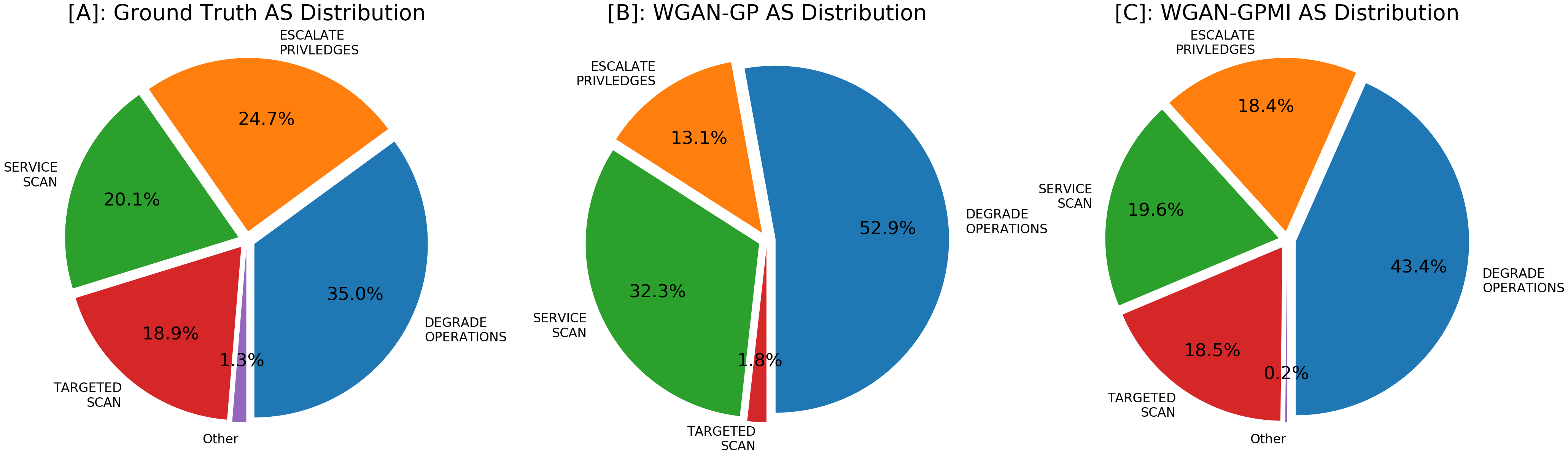}
    \caption{Distribution of Attack Stages (AS) on target IP 10.0.0.22 from CPTC'17. Note that the WGAN-GPMI model results [C] have a much closer probability distribution to the ground truth data [A] then the WGAN-GP Model [B]. \vspace*{-10pt}}
    \label{fig:killchain_results}
\end{figure*}

Finally, a means to identify the type of behavior associated with the additional output modes captured would provide contextual information to what type of network behaviors are most recoverable from data driven models such as GANs. To accomplish this Alert Signatures are mapped to one of the 12 attack stages provide in Table \ref{tab:attack_stages}. Figure \ref{fig:killchain_results} shows the attack stage coverage within the ground truth data as well as those generated from WGAN-GP and WGAN-GPMI for the CPTC'17 dataset. Note that the WGAN-GPMI model shows almost identical attack stage distribution as the ground truth data, exhibiting improvement over the WGAN-GP case. Specifically, it synthesized alerts pertaining to the Targeted Scanning stage with probability very close to the ground truth distribution. Meanwhile, the standard WGAN-GP model could not capture this output mode with probability greater than $1.8\%$, leaving a large gap in the generated data sample. 

\begin{figure}[!htbp]
    \centering
    \includegraphics[width=\columnwidth]{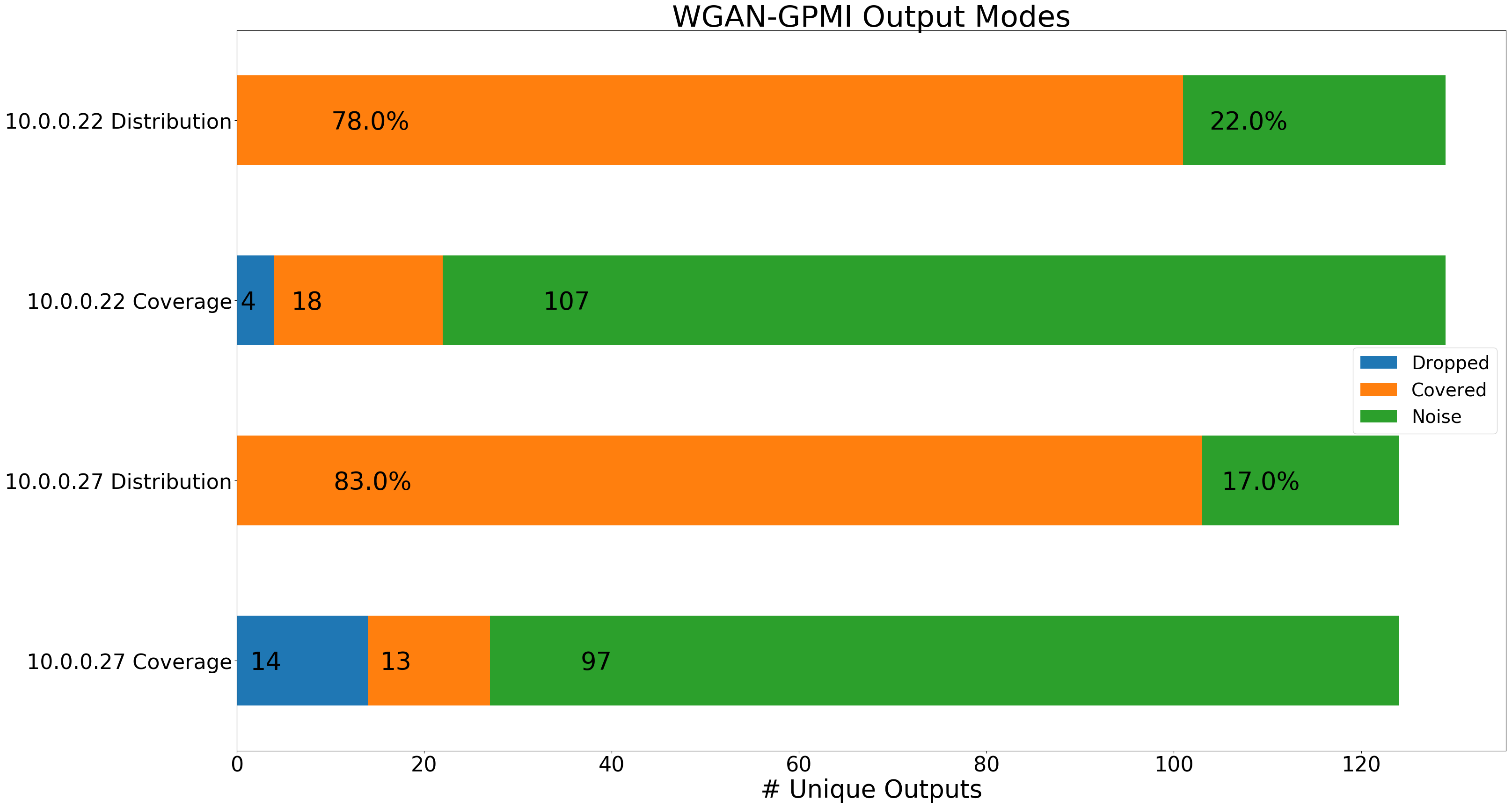}
    \caption{CPTC'17 Target IPs 10.0.0.22 and 10.0.0.23: The WGAN-GPMI model features less mode dropping than the WGAN-GP model, however the amount of probability mass assigned to noisy samples also increases. \vspace*{-10pt}}
    \label{fig:overall_drop}
\end{figure}

Figure \ref{fig:overall_drop} shows these results for the two target IPs as a series of bars. The bars marked "Coverage" show the number of unique alert combinations (modes) that fall into each category. The bars marked "Distribution" show the percentage of alerts from the generated distribution belonging to each category. 

First, note that a large percentage of alerts were covered for both target IP cases, whereas 17$\sim$20\% alerts with modes not seen in the ground truth data. A detailed look at the results reveal differences in the two target IP cases. Target 10.0.0.22 shows superior coverage with only 4 output modes being dropped while adding a larger number of novel modes (though the percentage of alerts is still a minority). On the other hand, Target 10.0.0.27 has less noisy modes and alerts but still drops 14 modes from the ground truth data. It is possible that these dropped modes represent samples which have an extremely low probability of occurring; so much so that the mutual information constraint is insufficient to encourage the generation of these values. Further supporting this is the fact that even with less than half of the total output modes covered there is still an 83\% chance that the outputs from this model do exist in the ground truth distribution. 

\section{Concluding Remarks}
\label{sec:conclusion}

This research showed the promise of using GAN to recreate target based cyber-alert data from known malicious alert datasets. Additionally, intra-alert dependencies are shown to be easily revealed through the usage of histogram intersection score between ground truth and generated alerts. Even when histogram complexity rises and intersection scores fall below 60\% the feature dependencies identified by GAN are shown to be accurate through conditional and joint entropy computation. 

The ability to identify intra-alert feature dependencies opens the door to many practical uses and future work. One such example would be increasing the utility and responsiveness of network intrusion prevention systems by updating detection rule-sets automatically and continuously on a per system basis; all based off the traffic previously seen. Furthermore, the analysis methods provided here are applicable to any discrete dataset, not just cyber-security alert data. 

% Discuss guiding generator behavior?

Finally, future experimentation with a GAN model configured to capture temporal dependencies through the usage of LSTM or CNN architectures would be of great benefit. Such a model would have the ability to find long term patterns in network behavior for a given machine, help to build attacker models, rather than identifying attack stage based off a single Alert Signature, and allow for complex multistep attacks to be learned. 

\ifCLASSOPTIONcaptionsoff
  \newpage
\fi

\newpage

\bibliographystyle{IEEEtran}
\bibliography{main}

\end{document}